\documentclass[sigconf,authorversion,nonacm]{acmart}
\AtBeginDocument{%
  \providecommand\BibTeX{{%
    \normalfont B\kern-0.5em{\scshape i\kern-0.25em b}\kern-0.8em\TeX}}}

\acmConference[Woodstock '18]{Woodstock '18: ACM Symposium on Neural Gaze Detection}{June 03--05, 2018}{Woodstock, NY}
\acmBooktitle{Woodstock '18: ACM Symposium on Neural Gaze Detection,June 03--05, 2018, Woodstock, NY}
\renewcommand\footnotetextcopyrightpermission[1]{}

\begin{document}

\title[Towards Designing Spatial Robots that are Architecturally Motivated]{Towards Designing Spatial Robots \\that are Architecturally Motivated}

\author{Binh Vinh Duc Nguyen}
\email{alex.nguyen@kuleuven.be}
\orcid{0000-0001-5026-474X}
\affiliation{
  \institution{Research[x]Design, \break Department of Architecture, KU Leuven}
  \streetaddress{Kasteelpark Arenberg 1 - box 2431}
  \city{Leuven}
  \country{Belgium}
  \postcode{3001}
}

\author{Andrew Vande Moere}
\email{andrew.vandemoere@kuleuven.be}
\orcid{0000-0002-0085-4941}
\affiliation{
  \institution{Research[x]Design, \break Department of Architecture, KU Leuven}
  \streetaddress{Kasteelpark Arenberg 1 - box 2431}
  \city{Leuven}
  \country{Belgium}
  \postcode{3001}
}


\begin{abstract}
While robots are increasingly integrated into the built environment, little is known how their qualities can meaningfully influence our spaces to facilitate enjoyable and agreeable interaction, rather than robotic settings that are driven by functional goals. Motivated by the premise that future robots should be aware of architectural sensitivities, we developed a set of exploratory studies that combine methods from both architectural and interaction design. While we empirically discovered that dynamically moving spatial elements, which we coin as spatial robots, can indeed create unique life-sized affordances that encourage or resist human activities, we also encountered many unforeseen design challenges originated from how ordinary users and experts perceived spatial robots. This discussion thus could inform similar design studies in the areas of human-building architecture (HBI) or responsive and interactive architecture. 
\end{abstract}



\begin{teaserfigure}
  \includegraphics[width=\textwidth]{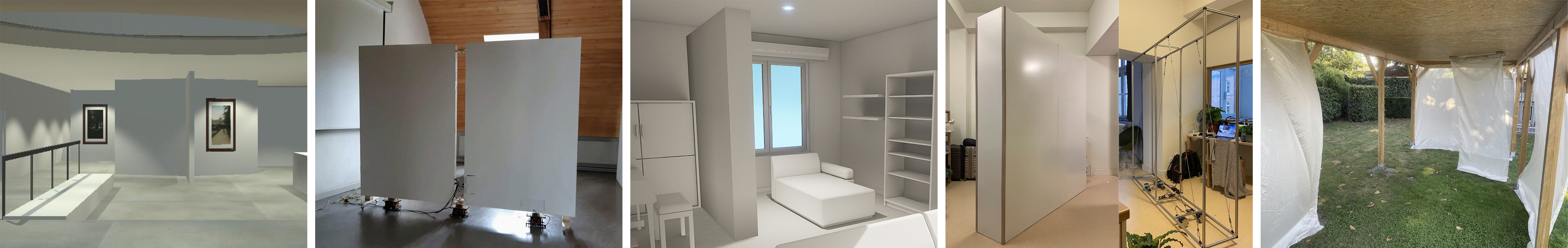}
  \caption{The five studies of architecturally-motivated spatial robots that we have deployed, which explored space-dividing elements in different expressions, medium and contexts. From left to right: (1) a set of moving, connected panels in a virtual museum space \cite{Nguyen2020}; (2) an autonomous, robotic folding panel in a real university space \cite{Nguyen2020}; (3) a moving wall in the semi-immersive cross-reality simulations of people's real homes \cite{Nguyen2021}; (4) an autonomous, robotic moving wall in people's real apartments; and (5) an autonomous, robotic moving curtain in a semi-outdoor, semi-public space. Studies (1) and (2) were reported in a publication . Study (3) will be presented in the upcoming HRI'21 conference . Study (4) is currently a pilot study and will be deployed soon.}
  \label{fig:teaser}
\end{teaserfigure}

\maketitle

\section{Introduction}

The integration of robotic technology into the built environment is an emerging trend promised to widen the scope of how our living space is experienced \cite{Bolbroe2016}. Originated as a architectural vision from as early as the 1960s, this integration should enable spaces to dynamically adapt to the functional needs of their occupants \cite{Eastman1971, Pask1969}, to become a "\textit{fluid, vibrating backdrop for the varied and constantly changing modes of life}" \cite{Zuk1970}. As many relevant technological innovations like digital manufacturing, pervasive sensing, mechanical actuation and artificial intelligence are becoming increasingly accessible and affordable, groundbreaking opportunities are landing in the hands of architectural practice to materialise this vision towards purposeful goals. From this development, there emerges the design of physically moving elements aiming to influence the spatial dimensions of our built environment, what we coin as \textit{spatial robots}.

In the discipline of responsive and interactive architecture, visionary thinkers already proposed that spatial robots can affect the behaviour, thoughts or emotional states of occupants in ways that are perhaps more compelling than ‘static’ architecture \cite{Meagher2015}. Towards this goal, they suggested that those robots should interact with occupants in a conversational dialog \cite{Oosterhuis2012, Schnadelbach2016} through poetic forms of spatial expression \cite{Meagher2015} that are reminiscent to attitudes \cite{Achten2013}. Yet, despite technological explorations in relevant sub-domains like architectural robotics \cite{GrossRobotics2012} or human-building interaction (HBI) \cite{AlaviHBI2019,Wiberg2020}, we still do not know yet how to design spatial robots that meaningfully influence the experience of occupants as much or even more than its "static" architectural counterpart, towards altruistic goals such as to benefit people health and well-being \cite{Roessler2012} or quality of life \cite{Steemers2020}.

Inspired by these ideas, we propose that in order to create beneficial experiences with occupants, the design of spatial robots, such as their materiality or behavior, should be motivated by design principles from architectural theory. As such, we wish to attain a better understanding of the architectural dimensions of purposeful human-robot interactions in our daily life. We thus deployed a set of exploratory studies that involved ordinary people experiencing spatial robots in a range of real-world situational contexts, as shown in Figure \ref{fig:teaser}. To adequately capture the experience of participants, we used the mixed-method research approach that deployed methods originating from two relevant disciplines: architectural and interaction design. 

Although our studies confirmed the hypothesis that people can interpret the spatial impact of architectural robots, they also revealed many unforeseen challenges that highlighted the unfamiliarity of ordinary users to spatial robots and the discrepancies between experts. We believe that this discussion is not only useful for studies in HBI and architectural robotics in general, but can also be applicable for other cross-disciplinary HRI research that aims to define the behavior of robots using fuzzy designerly approaches.

\section{Design Process}

We implemented each study through an iterative design process towards the research objectives.
It is naturally not easy to find available contexts that allows experiments in architectural scales to take place, especially in private spaces such as offices or homes. In order to create the "sense of belonging", the design of spatial robots should also architecturally contribute to these contexts using expressions of materiality or behavioral patterns. As such, each of our study involved several pilot tests to achieve the suitable design decisions before the actual deployment.

For example, Study 1, which took place in a museum space, implemented a set of interactive, three-winged connected panels to guide participants through the exhibition with crisscross movements, encouraging their curiosity with new discoveries every new corner \cite{Nguyen2020}. Study 2, which took place in an open-plan university space, deployed an autonomous, folding panel to effectively transform an open, public space into semi-closed, semi-public sub-spaces \cite{Nguyen2020}. Study 3, 4 and 5 specifically aimed to maintain safety during the COVID-19 pandemic, by involving participants experiencing a responsive, moving wall in their own homes without physical contact via a semi-immersive cross-reality simulation \cite{Nguyen2021}, or interacting with a robotic moving curtain in an outdoor space to reduce potential risks.

\section{Research Challenges}

\subsection{From the Perspective of Ordinary Users}

Our empirical findings show that ordinary users generally perceived spatial robots in a distinctly different manner compared to "static" architecture, in at least three ways:

\subsubsection{Necessity}

The purpose of spatial robots are difficult for some of our participants to comprehend. They expressed that humans can freely move from one space to another, or change the configurations of each space at will. As such, they believed that the needs for architecture to physically move were only necessary in spaces that need to host a multitude of spatial usage scenarios. These evidences show that architecture is still generally perceived as a passive, static, contextual background for our daily lives. The motivation for flexible architecture is mostly thought to be functional, as shown in the available smart-home market. Meanwhile, the experiential dimension of spatial robots is still largely unexplored.

\subsubsection{Applicability}

Some participants questioned the feasibility to integrate spatial robots in their daily living space, mentioning various safety or privacy concerns. They expressed that architecture is structural, weight-bearing and thus should be strong and robust to withstand the movements or required loads of daily activities.
Some concerned about the risk of including data collecting equipment in their private spaces, as it is a inevitable component to make spatial robots "aware" of the context they should "respond" to. 
The design of spatial robots, therefore, should take in considerations these aspects ethically, to make them become safe, reliable companions. A potential approach is to recognize the context from an architectural perspective, such as to collect noise level rather than words, or to sense lighting level caused by the movements of occupants rather than recording their positions.

\subsubsection{Practicality}

Differently from "static" architecture, spatial robots have the unique power to communicate to the participants and move between their levels of attention using location and motion. Our results suggest that a curtain next to a window conveyed a different meaning than a curtain in a middle of an empty room, or different locations of a wall could facilitate or hinder several architectural atmosphere to occur \cite{Nguyen2021}. While the participants rarely noticed an immobile spatial division, most of them disrupted their activity once it started 'waving' autonomously or moving closely to their location. 
As such, we noticed that the architectural experiences were amplified with the present of spatial robots. With "static" architecture, these experiences are usually ambient, in the periphery of occupants attention \cite{Rooney2017}
%, unless the space is spectacular or there are elements that disturb their activity. 
Spatial robots, therefore, have the potentials to generate a new set of expressions with unique affordances. Yet, the challenge is now how to design or research such robots that influence our experience through both focal object perception and ambient spatial perception.

\subsection{From the Perspective of Experts}

While the theoretical differences between architectural and interaction design have been discussed extensively in the discipline of HBI \cite{Lundgaard2019, Kirsh2019}, we encountered practical challenges with this convergence of disciplines in at least three ways:

\subsubsection{Terminology}

The selection of terminology while studying spatial robots could highlight the discrepancies between architectural and interaction design. 
For example, popular functional disruptive terms in architecture such as "\textit{responsive, interactive}" might not be suitable from the viewpoint of HCI, as they describe robots with intention rather than mediating testbeds that follow instructions. Material terms such as "\textit{wall}" can be related to a spatial divider that affects the privacy, acoustic, or visual qualities in user studies. Yet, when this "wall" is not robust, weight bearing or even static, it is not adequately a "wall" in architecture.
This challenge originates from the fact that while architects are accustomed to fuzzy terms that aim to convey designerly intentions, HCI researchers usually carefully analyse the affordances of those terms since they might affect the reliability of the experience.
As designing spatial robots is cross-disciplinary, it requires the selection of "boundary terms" that respects both architectural and interaction design, in order to facilitate a smooth collaboration between experts. 

\subsubsection{Methodology}

Studying spatial robots require a methodology that can adequately capture the experience, while reconciling the different viewpoints within architectural and interaction design \cite{Kirsh2019}.
An effective methodology that we deployed is to implement a range of methods that captured spatial experience both qualitatively (i.e. with video recording, observations or semi-structured interviews) and quantitatively (i.e. with digital logging or questionnaires). However, integrating these methodology is challenging because it requires a balance between both positivist and constructivist perspectives \cite{Martina2010}. For example, positivist scientists might question the generalisability of such studies because architectural experience is naturally subjective, individual. Meanwhile, constructivist scientists might disagree with the causal relationships between architectural configuration and human experience, since they are too objective to reflect the personal, embodied experience as in architectural phenomenology \cite{lefebvre1992, merleau1962}.
A solution for this is perhaps to develop the behavioral frameworks of spatial robots as a collection of design patterns \cite{alexander1977, Kahn2008}, i.e. non-descriptive solutions that still require some mediate levels of interpretation to fit the design situation at hand, rather than concrete knowledge.

\subsubsection{Materiality}

As shown in Figure \ref{fig:teaser}, our exploration of spatial robots spanned through several dimensions of materiality, such as spatial contexts (i.e. private vs public space), medium (i.e. virtual vs physical reality), or expression (i.e. curtain vs wall). 
While our findings evidenced that these dimensions affected the experience of participants to some extents, the design choices that we made were determined by the context rather than scientifically driven.
As such, the approach to study spatial robots can come from the architectural, designerly perspective, which is essentially contextual. Yet, to sufficiently cover the ground of influential factors, there might require some metrics or criteria that are independent from the design process. Some suggestions could be the dimensions of architectural affordances (i.e. functionality, visuality, connectivity, etc.) or levels of responsiveness of the spatial robots (i.e. fully controlled, semi-controlled, autonomous).

\section{Conclusion}

With a body of exploratory studies, we identified the design challenges that address how the study of spatial robots can be developed further to benefit people health and wellbeing or quality of life. We unpacked these challenges based on how ordinary users perceived spatial robots through its necessity, applicability and practically; and how experts from architectural and interaction design saw their discrepancies in terminology, methodology and materiality. Through these challenges, we highlighted the situational, contextual aspects when designing spatial robots. We thus proposed that the integration of architectural knowledge within HRI is necessary to convince designers and architects to use these technologies. This integration has the potentials to facilitate human-building interactions that are enjoyable and agreeable, instead of designing buildings as robotic configurations driven by functional goals.

\begin{acks}
The research reported in this paper is funded by grant CELSA/18/020 titled “Purposefully Controlling Mediated Architecture”.
\end{acks}

\bibliographystyle{ACM-Reference-Format}
\balance 
\bibliography{sample-base}

\end{document}